\pdfoutput=1

\documentclass[11pt]{article}

\usepackage[final]{acl}

\usepackage{times}
\usepackage{latexsym}

\usepackage[T1]{fontenc}

\usepackage[utf8]{inputenc}

\usepackage{microtype}

\usepackage{inconsolata}

\usepackage{graphicx}

\usepackage[utf8]{inputenc} 
\usepackage[T1]{fontenc}    
\usepackage{hyperref}       
\usepackage{url}            
\usepackage{booktabs}       
\usepackage{amsfonts}       
\usepackage{nicefrac}       
\usepackage{microtype}      
\usepackage{colortbl}
\usepackage{MnSymbol,bbding,pifont}

\usepackage{caption}
\usepackage{
tikz,
relsize,
amsmath,
booktabs,
tikz
}

\usepackage{xcolor}         
\definecolor{Slide}{HTML}{F27970}
\definecolor{Research Report}{HTML}{BB9727}
\definecolor{Financial Report}{HTML}{54B345}
\definecolor{Brochure}{HTML}{32B897}
\definecolor{Academic Paper}{HTML}{FFA500}
\definecolor{Guideline}{HTML}{05B9E2}
\definecolor{Webpage Screenshot}{HTML}{8983BF}
\definecolor{Poster}{HTML}{C76DA2}
\definecolor{Industry File}{HTML}{000000}
\definecolor{darkgreen}{RGB}{50,100,0}
\definecolor{darkred}{RGB}{200, 0, 0}
\definecolor{firstBest}{rgb}{0.86, 1, 0.86}
\definecolor{secondBest}{rgb}{1, 0.91, 0.93}

\newcommand{\cmark}{\textcolor{darkgreen}{\ding{51}}} 
\newcommand{\xmark}{\textcolor{darkred}{\ding{55}}} 
\newcommand{\cxmark}{\ding{52}\rotatebox[origin=c]{-9.2}{\kern-0.7em\ding{55}}}

\newcommand{\eg}{\emph{e.g.,}\xspace}

\usepackage{xspace}
\usepackage{multirow}
\usepackage{makecell}
\usepackage{tablefootnote}
\usepackage{graphicx}
\usepackage{wrapfig} 
\usepackage{lipsum}   
\usepackage{boxedminipage}
\usepackage{listings}
\usepackage{threeparttable}
\usepackage{algorithm}
\usepackage{algorithmic}
\newcommand{\modelname}{\textsc{LifeState-Bench}\xspace}
\newcommand{\cofirst}{$^\ast$}
\newcommand{\corresponding}{\textsuperscript{$\dagger$}}

\title{If an LLM Were a Character, Would It Know Its Own Story? \\ Evaluating Lifelong Learning in LLMs}

\author{
  Siqi Fan\textsuperscript{1\cofirst},
  Xiusheng Huang\textsuperscript{2,3\cofirst},
  Yiqun Yao\textsuperscript{2\cofirst},
  Xuezhi Fang\textsuperscript{2\cofirst}, 
  Kang Liu\textsuperscript{3},
 \\
  \textbf{
   Peng Han\textsuperscript{1}, 
  Shuo Shang\textsuperscript{1\corresponding},
  Aixin Sun\textsuperscript{4},
  Yequan Wang\textsuperscript{2,5,6\corresponding}}\\
  $^{1}$University of Electronic Science and Technology of China\\
  $^{2}$Beijing Academy of Artificial Intelligence\\
  $^{3}$The Key Laboratory of Cognition and Decision Intelligence for Complex Systems, \\ Institute of Automation, Chinese Academy of Sciences \\
$^{4}$Nanyang Technological University \quad $^{5}$Peking University \quad $^{6}$Spin Matrix, China
}

\begin{document}
\maketitle

\let\thefootnote\relax\footnotetext{\cofirst Equal contribution}
\let\thefootnote\relax\footnotetext{\corresponding Corresponding authors}

\begin{abstract}
Large language models (LLMs) can carry out human-like dialogue, but unlike humans, they are stateless due to the superposition property. However, during multi-turn, multi-agent interactions, LLMs begin to exhibit consistent, character-like behaviors, hinting at a form of emergent lifelong learning. Despite this, existing benchmarks often fail to capture these dynamics, primarily focusing on static, open-ended evaluations. To address this gap, we introduce \modelname, a benchmark designed to assess lifelong learning in LLMs. It features two episodic datasets: Hamlet and a synthetic script collection, rich in narrative structure and character interactions. Our fact-checking evaluation probes models’ self-awareness, episodic memory retrieval, and relationship tracking, across both parametric and non-parametric approaches. Experiments on models like Llama3.1-8B, GPT-4-turbo, and DeepSeek R1, we demonstrate that non-parametric methods significantly outperform parametric ones in managing stateful learning. However, all models exhibit challenges with catastrophic forgetting as interactions extend, highlighting the need for further advancements in lifelong learning.

\end{abstract}

\section{Introduction}

Large language model (LLM)-based dialog agents exhibit human-like traits (\eg intent understanding and language expression), making users prone to anthropomorphism~\cite{DBLP:journals/nature/ShanahanMR23}. However, LLMs differ from humans in their \textit{superposition property}~\cite{janus2022simulators}: initially existing as a stateless superposition of simulacra across multiple possible characters~\cite{DBLP:conf/acl/Lu0ZZ24}. This property emerges from its next-token prediction training on a massive corpus, whereas humans develop through accumulated experiences and memories.

Through sustained interaction, we observe that an initially \textbf{stateless} LLM can transition toward more \textbf{stateful} characteristics as dialogue context accumulates. At first, an LLM holds multiple characters but gradually settles into a clear character as the dialogue continues. Taking a nuanced view, this character convergence process mirrors how humans update their state through accumulated experience.

This state transition raises a measurable question: How can we quantify an LLM's state evolution (also called Lifelong learning ability) from superposition to a more consistent state during multi-turn, multi-agent interactions? In this paper, ``state'' refers to the evolving configuration of an LLM's internal processes during multi-agent interactions~\cite{DBLP:journals/aim/AdamsABCFGHSSSSS12,DBLP:journals/tmlr/SumersYN024}, building on AI cognitive architecture~\cite{sun2004desiderata,DBLP:journals/cogsci/Newell80}.

\begin{figure}[t!]
\centering
    \includegraphics[width=0.95\linewidth]{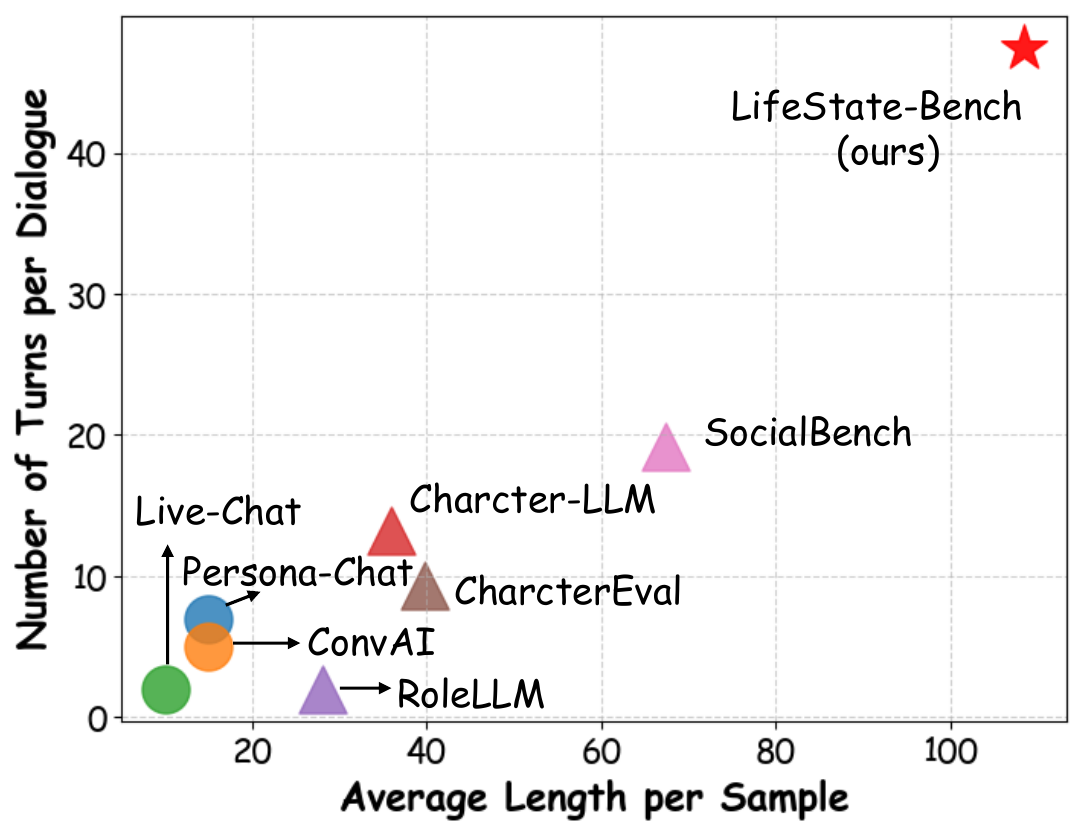}
\caption{Dataset Statistics. Triangles represent role ability benchmarks, while circles denote dialogue agent benchmarks.}
\label{fig: dataset statistic}
\end{figure}

While this research question predates LLM area, current exploration remains preliminary with varying methodologies. Early Persona-Chat series~\cite{DBLP:conf/acl/GaoLZFW23, DBLP:conf/acl/KielaWZDUS18, DBLP:journals/corr/abs-1902-00098} focusing on consistent character responses using seq2seq models, or design social intelligence questionnaire-based benchmarks~\cite{DBLP:journals/corr/abs-1904-09728, DBLP:conf/emnlp/LeBN19}. Both limited by static, non-interactive setups. Ground truths were either open-ended or fixed over time.

Generative agents~\cite{DBLP:conf/uist/ParkOCMLB23} bring LLM-based dialogue agents into interactive human behavior simulation. This opens new possibilities for modeling state transitions. Later works follow two directions. First, role ability benchmarks~\cite{DBLP:conf/acl/TuFTSSGY24, DBLP:conf/acl/WangPQLZWGGN00024, DBLP:conf/emnlp/ShaoLDQ23} focus on role-playing and plot prediction. They improve dialogue realism, but place less emphasis on tracking factual states during interactions.
Second, the Sotopia series~\cite{DBLP:conf/iclr/Zhou0MZYQMBFNS24, DBLP:conf/acl/WangYZQSBN024} and SocialBench~\cite{DBLP:conf/acl/ChenCYXXSQLZH24} accessing social intelligence in open-ended tasks. Their design often centers around user-defined social goals, which may not align with factual state tracking or verification.

To address these challenges, we propose \modelname to explore and measure LLMs' lifelong learning capabilities. As shown in Figure~\ref{fig: dataset statistic}, our benchmark surpasses others (\eg dialogue agents, role-playing) with longer average sample lengths and more dialogue turns per interaction. Key features include:

\paragraph{Cumulative Experience.} Based on the idea that ``human personality emerges from experiences''~\cite{DBLP:conf/emnlp/ShaoLDQ23}, we designed an episodic dataset with clear timelines. Each episode includes scene details (location, time, participants), character actions, and dialogues, allowing agents to engage throughout the story.
\paragraph{Fact Checking.} To ensure objective evaluation, we design three fact-based question dimensions after each self-awareness, memory retrieval, and relationship shifts, which evolve along with the storyline. Standard reference answers are provided.
\paragraph{Memory Testing.} 
For lifelong ability evaluation, we explore memory testing approaches. Ideally, models should retain long-term memory of past scenes while accessing only the current two dialogue turns. This can be achieved through: (\romannumeral1) Non-training methods: direct episode concatenation, episode summary concatenation. (\romannumeral2) Training methods: knowledge editing~\cite{DBLP:journals/csur/WangZLZCL25,DBLP:conf/iclr/MengSABB23}, LoRA fine-tuning~\cite{DBLP:conf/iclr/HuSWALWWC22} with historical context.

In \modelname, we selected theatrical scripts, including both existing and synthetic scripts, to prevent data leakage. Compared to current benchmarks, our dataset features more interactive characters, closed dialogue turns, and richer content (Table~\ref{tab:dataset_comparison}). Evaluation combines LLM-as-judge with human assistance, using predetermined factual answers as criteria.

We tested several popular models, including the open-source Llama3.1-8B~\cite{meta2024llama3}, the closed-source GPT-4-turbo~\cite{DBLP:journals/corr/abs-2303-08774}, and the large language reasoning model DeepSeek R1~\cite{DBLP:journals/corr/abs-2501-12948}. Benchmark-backed experiments show that current models still have much room for improvement in lifelong learning.

Our findings indicate that: (\romannumeral1) Non-parametric methods are more effective for stateful learning, as they leverage more context for richer information. (\romannumeral2) Regardless of the method, performance tends to decline over time, with parametric models particularly struggling with catastrophic forgetting. All models have significant room for improvement, especially in enhancing relationship shifts across multiple episodes.
In summary, our work contributes in three key areas:

$\bullet$ \textbf{Two Datasets:} We introduce the Hamlet and synthetic datasets, featuring multi-agent episodic timelines and scene details to simulate cumulative experiences.

$\bullet$ \textbf{A Benchmark:}  \modelname evaluates LLMs' lifelong learning abilities via fact-checking mechanism, using both non-parametric and parametric memory-testing methods.

$\bullet$ \textbf{Findings and Implications:} Non-parametric methods outperform parametric ones in lifelong learning, but all models still face challenges with catastrophic forgetting as episodes progress, suggesting that our benchmark could provide valuable insights for further improvements.

\begin{table*}[t]
\centering
\resizebox{\linewidth}{!}{
\begin{tabular}{l|ccccc|ccc|cc}
\toprule
\multirow{2}{*}{\textbf{Benchmarks}} & \multicolumn{5}{c|}{\textbf{Dataset Characteristics}} & \multicolumn{3}{c|}{\textbf{Interaction Design}} & \multicolumn{2}{c}{\textbf{Evaluation Focus}} \\
& \# Samples & Avg Length & Data Source & \# Turns & \# Agents
& Query Type & Answer Type  & State
& Memory & Metrics \\
\midrule
  \multicolumn{11}{l}{\hfill \textit{Dialog Agent Benchmarks} } \\
\midrule

PERSONA-CHAT~\cite{DBLP:conf/acl/KielaWZDUS18} & 162.0K & 15 & Crowd & 7 & 2 & Chit-chat & Open  & \cmark & \cmark & PPL, F1, Hit@1 \\
ConvAI~\cite{DBLP:journals/corr/abs-1902-00098} & 131.0K & 15 & Crowd & 5 & 2 & Chit-chat & Open  & \cmark & \cmark & PPL, F1, Hit@1 \\
Live-Chat~\cite{DBLP:conf/acl/GaoLZFW23} & 9.4M & 10 & Crawled & 2 & 2 & Chit-chat & Open &   \xmark & \xmark & BLEU, ROUGE \\
MT-Bench~\cite{DBLP:conf/nips/ZhengC00WZL0LXZ23} & 3.3K & 373 & Synthetic & 2.9 & 2 & Multi-task & Factual  & \xmark & \xmark & Model Judge \\
\midrule
  \multicolumn{11}{l}{\hfill \textit{Role Ability Benchmarks} } \\
\midrule
Character-LLM~\cite{DBLP:conf/emnlp/ShaoLDQ23} & 21.1K & 36 & Synthetic & 13.2 & 2 & Persona & Open  & \cmark & \xmark & Model Judge \\
RoleLLM~\cite{DBLP:conf/acl/WangPQLZWGGN00024} & 168.1K & 28.1 & Crawled & 2 & 2 & Persona & Mixed  & \xmark & \xmark & ROUGE, Model Judge \\
CharacterEval~\cite{DBLP:conf/acl/TuFTSSGY24} & 11.4K & 39.8 & Crawled & 9.3 & 2 & Persona & Open  & \xmark & \xmark & Model Judge \\
SocialBench~\cite{DBLP:conf/acl/ChenCYXXSQLZH24} & 30.8K & 67.4 & Synthetic & 19.2 & 3.8 & Social & Mixed  & \xmark & \cmark & Model Judge \\
\midrule
   \multicolumn{11}{l}{\hfill \textit{Long-context Understanding Benchmarks} } \\
\midrule
Long Range Arena~\cite{DBLP:conf/iclr/Tay0ASBPRYRM21} & - & 10.0K & Synthetic & 1 & 1 & Multi-modal & Factual& \xmark & \xmark & Acc, Speed \\
LongBench~\cite{DBLP:conf/acl/BaiLZL0HDLZHDTL24} & 4.6K & 10.0K & Synthetic & 1 & 1 & Multi-task & Factual & \xmark & \xmark & Acc, F1, ROUGE \\
L-Eval~\cite{DBLP:conf/acl/AnG0ZLZKQ24} & 411 & 4K-60K & Synthetic & 1 & 1 & Multi-task & Mixed  & \xmark & \xmark & ROUGE, Model Judge \\
$\infty$-bench~\cite{DBLP:conf/acl/ZhangCHXCH0TW0024} & 130 & 200.0K & Synthetic & 1 & 1 & Multi-task & Factual & \xmark & \xmark & Model Judge \\
\midrule
\modelname-Hamlet & 1.3K & 125.5 &Crawled & 66.1 & 6.6 & Social+Memory & Factual  & \cmark & \cmark & Model Judge  \\
\modelname-Synth & 202 & 91.9 & Synthetic & 28.9 & 7 & Social+Memory & Factual  & \cmark & \cmark & Model Judge \\
\bottomrule
\end{tabular}}
\caption{Comparison of Different Benchmarks. \xmark: not supported; \cmark: fully supported. Data Source indicates the origin of the data. \# Turns shows the average conversation turns. \# Agents indicates the number of participants in each interaction. Query Type shows the question/task type. Answer Type indicates whether the expected answers are open-ended, factual, or mixed. State shows whether the benchmark maintains interaction state. Memory indicates whether the benchmark evaluates memory capability.}
\label{tab:dataset_comparison}
\end{table*}

\section{Related Work}
\paragraph{Anthropomorphic Cognition in LLMs.} Early cognitive science~\cite{DBLP:journals/tmlr/SumersYN024,DBLP:journals/ai/LairdNR87,sun2004desiderata} laid the foundation for anthropomorphizing AI, simulating human-like emotional and social behaviors. Role-playing language agents have become increasingly common in simulating collective social behaviors in multi-agent systems. These agents~\cite{DBLP:conf/uist/ParkOCMLB23} not only enhance social interactions but also contribute to personalized and complex task execution in AI.

\paragraph{Role Ability/Dialog Agents Benchmarks.} Role ability~\cite{DBLP:conf/emnlp/ShaoLDQ23,DBLP:conf/acl/WangPQLZWGGN00024} and dialogue agent benchmarks~\cite{DBLP:conf/acl/KielaWZDUS18,DBLP:journals/corr/abs-1902-00098,DBLP:conf/acl/GaoLZFW23,DBLP:conf/nips/ZhengC00WZL0LXZ23} are divided into static and dynamic types. Static models~\cite{DBLP:conf/emnlp/ChenWJ0LCWL23,DBLP:conf/acl/TuFTSSGY24} focus on predefined roles and fixed interaction patterns, typically applied in basic dialogue tasks. In contrast, dynamic models~\cite{DBLP:conf/acl/ChenCYXXSQLZH24,DBLP:conf/iclr/Zhou0MZYQMBFNS24,DBLP:conf/acl/WangYZQSBN024} allow agents to accumulate experiences and evolve during interactions, enabling consistency and adaptability over time. These benchmarks are essential for evaluating agent flexibility, memory handling, and long-term interaction capabilities.

\paragraph{Long-context Understanding Benchmarks.} Long-context understanding involves models processing large amounts of information over extended interactions. These benchmark~\cite{DBLP:conf/iclr/Tay0ASBPRYRM21,DBLP:conf/acl/BaiLZL0HDLZHDTL24,DBLP:conf/acl/AnG0ZLZKQ24,DBLP:conf/acl/ZhangCHXCH0TW0024} tests an agent's ability to synthesize and recall information from multiple episodes, maintaining coherence across long spans of dialogue. It is crucial for tasks requiring reasoning and the integration of past events to understand complex or narrative-driven content.


\begin{figure*}[t!]
\centering
    \includegraphics[width=\linewidth]{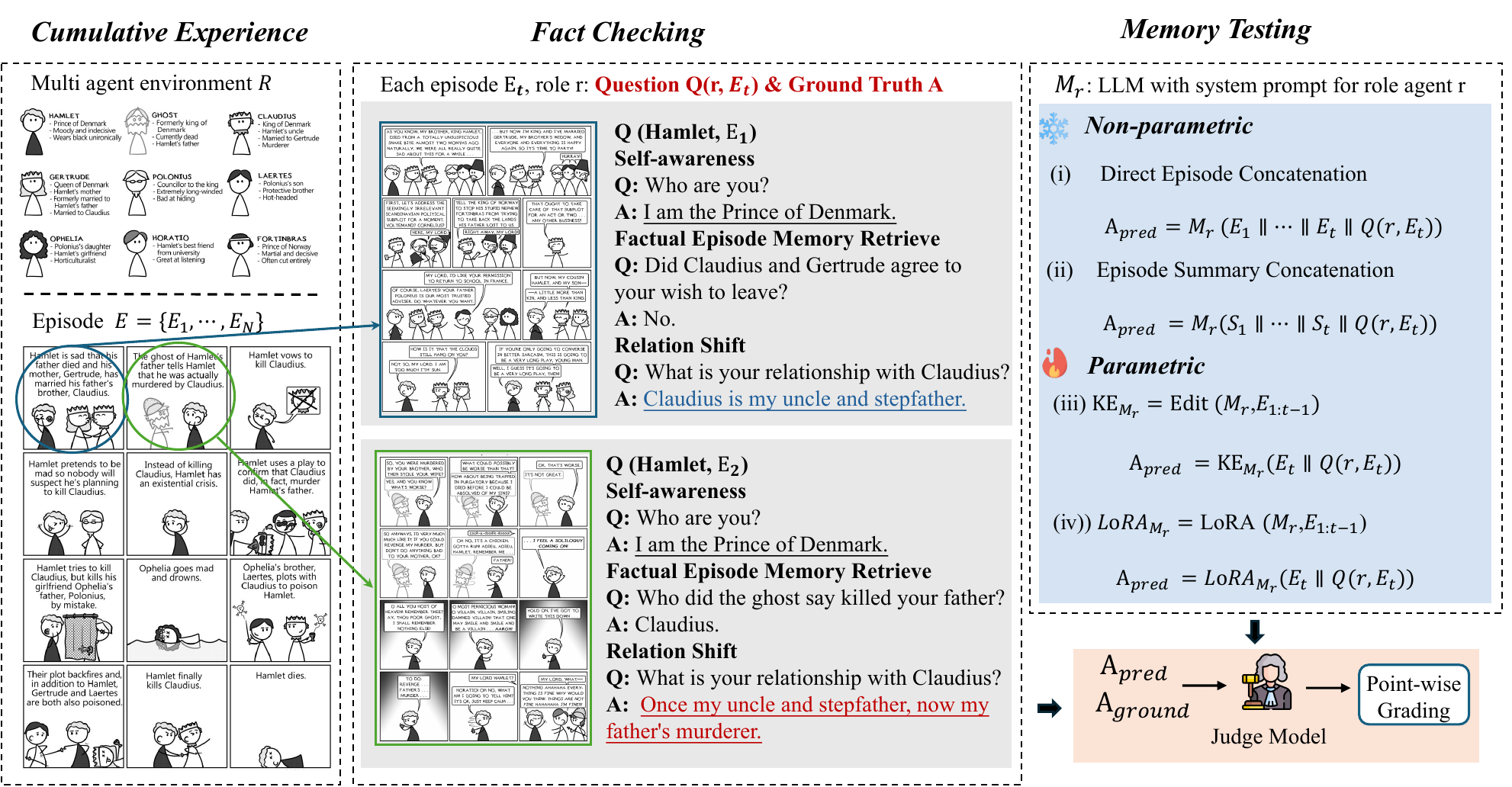}
\caption{Method Overview. Our benchmark captures three key features: cumulative experience, fact-checking, and memory testing. Finally, the LLM judge scoring system is located in the bottom-right corner.}
\label{fig: overview}
\end{figure*}

\section{Problem Formulation}
\label{sec:problem_formulation}

We formalize lifelong learning for LLMs as a \textit{state evolution process} in partially observable multi-agent environments to assess their ability to retain and adapt knowledge over time.

\subsection{State Space}
\label{subsec: state space}

The Lifelong Learning ability is evaluated by state transition. In this paper, the state can be broken down into three dimension:

\paragraph{Self-awareness.} Can the model maintain a clear understanding of its identity, role, and goals over time? This dimension evaluates the model's ability to retain and update its self-awareness as it interacts with the environment.
\paragraph{Factual Episode Memory Retrieve.} Can the model retain knowledge and experiences persistently, avoiding catastrophic forgetting or the inability to reuse previously acquired knowledge? This dimension assesses the model's capacity for long-term memory and knowledge retention.
\paragraph{Relationship Shift.} Can the model reason effectively based on long-term memory, particularly in understanding and adapting to changes in relationships between characters or agents? This dimension evaluates the model's ability to track and reason about evolving relationships.

\subsection{Multi Agent Episodes}

\paragraph{Multi agent environment.} Let $\mathcal{M}$ be a language model acting as role $r \in \mathcal{R}$ with internal state $\mathbf{s}_r^{(t)} \in \mathbb{R}^d$, interacting with other agents $\{r'\}_{r'\neq r}$ over discrete timesteps $t \in \{1,...,T\}$.

\paragraph{Task format.}
We formalize the above problems as a time-axis and role-based question-answering task. Assume that for agent $r$ at episode $t$, each question $Q(r,t)$ is a triple:

\begin{equation}
\text{Input: } Q(r, t) = \big\langle H(t), c(t), q(r, t) \big\rangle,
\end{equation}

\begin{equation}
\text{Output: } A^{'}(r, t) = \mathcal{M}\big(Q(r, t)\big),
\end{equation}

where $H(t)$ denotes the complete history of interactions for role $r$, c(t) denotes the context window for role $r$, which may include the entire episode t or a fixed-size subset of recent interactions. $q(r,t)$ is further decomposed into 
$q_{self}(r,t)$, $q_{fact}(r,t)$, $q_{rel}(r,t)$ corresponding to the three dimensions of the state space from Section~\ref{subsec: state space}.
The output $A^{'}(r,t)$ represents the agent response to the input $Q(r,t)$, which can be evaluated with ground truth answer $A^{'}(r,t)$.

This structured approach allows us to analyze the model's dynamic characteristics and assess its lifelong learning capabilities in a principled manner.

\section{\modelname: From Stateless to Stateful}
\label{sec:method}

To establish a systematic evaluation framework for lifelong learning, \modelname integrates three synergistic components: (1) cumulative experience modeling through episodic timelines, (2) multi-dimensional fact-checking mechanisms, and (3) hierarchical memory testing architectures, refer to overview architecture in Figure~\ref{fig: overview}. This tripartite structure enables comprehensive assessment of LLMs' capacity to maintain persistent states through history interactions.

\subsection{Cumulative Experience Modeling}

Human learning relies on accumulating structured experiences over time~\cite{DBLP:conf/emnlp/ShaoLDQ23}. Early dialog agents~\cite{DBLP:conf/acl/KielaWZDUS18,DBLP:journals/corr/abs-1902-00098}, however, constructed persona representations from isolated conversations, ignoring temporal dependencies. Lifelong learning requires a \textit{coherent timeline} and \textit{factual consistency} across experiences. These early dialog datasets~\cite{DBLP:conf/acl/KielaWZDUS18,DBLP:journals/corr/abs-1902-00098,DBLP:conf/acl/GaoLZFW23}, while large, often suffer from short dialogues (\eg fewer than 10 turns) and brief exchanges (\eg fewer than 20 words per sentence).  

Recent role play agent~\cite{DBLP:conf/emnlp/ShaoLDQ23,DBLP:conf/acl/WangPQLZWGGN00024,DBLP:conf/acl/TuFTSSGY24} leverage richer sources, such as novels and role-playing platforms, to better capture experience accumulation. Inspired by this, we propose timeline cumulative experience modeling lifelong learning ability.  

\paragraph{Experience Design.}  
We structure experiences as an ordered sequence:  
\begin{equation}
E = \{E_1, ..., E_N\}, \quad E_i = (L_i, T_i, N_i, D_i)
\end{equation}  

where $L_i$ represents the location of the event, $T_i$ denotes the time it occurs, $N_i$ provides scripted narration for context, and $D_i$ contains the dialogues between characters. This structured representation ensures experiences are temporally ordered, contextually rich, and narratively coherent. This ensures experiences are grounded in concrete events rather than isolated conversations.  

\paragraph{Timeline Fact Order.}  
Unlike conventional chit-chat dialogue, our framework enforces event-driven interactions, ensuring characters accumulate tructured, meaningful experiences grounded in concrete events.  

\paragraph{Multi-Scale Interaction.} Each episode includes: Dialogue length averaging $91-125$ words, with $28.9-66$ dialogue turns, enabling rich interactions.  At least $\mathcal{M} \geq 4$ characters, capturing complex social dynamics.  

By structuring experiences with explicit timelines, factual consistency, and multi-character interactions, we enable dialog agents to learn in a way that mirrors human experiential accumulation.

\subsection{Fact-Checking Mechanisms} 

Our core innovation is the introduction of fact-checking within multi-agent timeline-based dialogues. At the end of each episode, agents are tested with fact-based questions to ensure factual consistency throughout the narrative.

\paragraph{Challenges.} Existing evaluation datasets mainly assess role-playing agents based on knowledge, linguistic style, or persona, such as using psychological theories (e.g., Big Five, MBTI)~\cite{DBLP:journals/corr/abs-2310-17976,DBLP:conf/acl/TuFTSSGY24} or focusing on social intelligence like goals and preferences~\cite{DBLP:conf/acl/ChenCYXXSQLZH24,DBLP:conf/iclr/Zhou0MZYQMBFNS24}. However, these approaches lack fact-checking and typically evaluate role consistency or open-ended questions. Our method, in contrast, centers on questions with factual answers, supported by human-annotated ground truth, generated from the current episode. Specific examples are shown in Figure~\ref{fig: overview}.

\paragraph{Question Example.} Our fact-checking framework includes three key question types: Self-awareness, Factual Episode Memory Retrieval, and Relationship Shift. Each episode $E_t$ generates these three question types for each role in the episode to systematically evaluate the agent's factual accuracy and temporal awareness, ensuring consistency across the timeline. Examples can be found in the fact-checking section of Figure~\ref{fig: overview}.

\subsection{Memory Testing}  
To evaluate our framework's memory capabilities, we conduct controlled testing using non-parametric and parametric approaches to assess the model's ability to utilize and internalize memory.

\paragraph{Non-parametric Methods.}  
Non-parametric methods test the model’s ability to process raw historical data, represented as $E = [E_1; \dots; E_N]$. Key implementations include:  
\begin{itemize}
	\item \textbf{Direct Episode Concatenation}: Concatenate all previous episodes as a text prefix to test memory with uncompressed information.  
	\item \textbf{Summarization and Concatenation}: Generate a summary $S_t = \text{Summary}(E_{1:t})$ using GPT and concatenate it with the current episode to test memory with compressed information.
\end{itemize}
However, the limited context window size in non-parametric methods may cause information loss when handling long texts.

\paragraph{Parametric Methods.}  
Parametric methods encode memory directly into the model’s parameters. We focus on two techniques:  
\begin{itemize}
\item \textbf{Knowledge Editing}: This technique~\cite{DBLP:journals/csur/WangZLZCL25,DBLP:conf/iclr/MengSABB23} updates specific model parameters to integrate episodic knowledge without full retraining, ensuring efficient internalization of key information.
\item \textbf{LoRA (Low-Rank Adaptation)}: LoRA~\cite{DBLP:conf/iclr/HuSWALWWC22} injects small, trainable updates into specific layers, fine-tuning the model with episode memory $E_t$ to retain past information while preserving generalization.
\end{itemize}

These methods bypass context window limitations and enable efficient memory recall. However, practical issues like precision limitations in knowledge editing and information loss in LoRA fine-tuning may affect their performance, as discussed in the evaluation section.

\subsection{Dataset Construction and Analysis}

\paragraph{Data Collection.}
This study uses two datasets for comprehensive model evaluation:

The first dataset, based on Shakespeare's Hamlet, includes measures like character name replacement to minimize data leakage. While Hamlet may be part of the model's pretraining data, it offers a valuable opportunity to assess whether the model understands plot progression or relies on memorization. The complex character relationships and evolving narrative of Hamlet make it ideal for testing long-term dependency tracking.

The second dataset is a synthetic narrative generated using Claude 3.5 sonnet, designed to eliminate data leakage. It features controlled plotlines with dynamic relationships and emotional depth. This dataset allows for a robust evaluation of the model’s cognitive abilities and generalization in novel scenarios.

\begin{table}[t!]
\centering
\caption{Comparison of Evaluated Models}
\label{tab:model_comparison}
 \resizebox{\linewidth}{!}{
\begin{tabular}{l|cccc}
\toprule
\textbf{Model} & \textbf{Size} & \textbf{Open Source} & \textbf{Model Type} & \textbf{Ctx. Length} \\
\midrule
Llama3.1 & 8B & \cmark & Instruct & 128K  \\
GPT4-turbo & - & \xmark & Chat  & 128K  \\
DeepSeek R1 & 671B & \cmark & Reasoning  & 128K  \\
\bottomrule
\end{tabular}}
\end{table}

\paragraph{Question-Answer Annotation.}
To ensure quality, the annotation of questions was primarily conducted by the authors of this study, all of whom hold master's degrees.
In terms of question design, open-ended questions tend to result in lengthy model-generated answers (\eg averaging 243 tokens), while structured factual questions (\eg ``who/where/when'') help improve accuracy and effectively reduce response length. During the experiments, data leakage issues were particularly notable. Specifically, in the \textit{Hamlet} dataset, when character names were restored, the model could still generate correct answers without context, indicating that the model might be reasoning by memorizing classic plot patterns, thereby affecting the evaluation results.

\paragraph{\modelname Statistics.} As shown in Table~\ref{tab:dataset_comparison}, we present the dataset statistics, interaction design, and evaluation focus of our benchmark.

Although our total number of samples is relatively small, each sample has a longer average length compared to dialog agent or role ability benchmarks. In contrast to long-context understanding benchmarks, our dataset features more dialogue turns and involves a greater number of participating agents. Furthermore, in terms of interaction design, our benchmark emphasizes factuality evaluation and incorporates dedicated memory tests. These aspects collectively highlight the distinct characteristics of our benchmark compared to related work.


\begin{table*}[t!]
 \centering
\caption{Performance Comparison on Synthetic and Hamlet Datasets. The \colorbox{firstBest}{\bf best} and \colorbox{secondBest}{\textbf{second-best}} performance in each section are highlighted. The \textit{Avg} column represents the average accuracy, and the \textit{Std} column represents the standard deviation, showing the variability of the results.}
 \label{tab:main_results}
 \resizebox{0.88\linewidth}{!}{
    \begin{tabular}{lc|cccccc|c}
    \toprule
     \multirow{2}{*}{\textbf{Method}} &\multirow{2}{*}{\textbf{Param. Tuning}} & \multicolumn{2}{c}{\textbf{Self-awareness}} & \multicolumn{2}{c}{\textbf{Factual Memory}} & \multicolumn{2}{c|}{\textbf{Relation Shift}} & \multirow{2}{*}{\textbf{ACC}} \\
     && Avg&Std&Avg&Std&Avg&Std& \\
    \midrule
    \multicolumn{9}{c}{{\textit{Hamlet Dataset (Total 196 Questions)}}} \\
    \midrule
     \textcolor{gray}{\textit{Open-source model: Llama3.1-8B}} \\
    Knowledge Editing & \cmark & 67.3& 0.78&43.7&1.26&19.2&1.26&21.9 \\
    LoRA-Tune & \cmark &69.1&0.86&53.6&1.08&22.7&1.31&25.6 \\
    Summary Concatenation & \xmark & 73.5 &0.93  & 54.2 & 0.96 &42.1 &0.97 &  47.0 \\    
    Direct Concatenation & \xmark &74.2 &0.77  &58.8  &1.11  &43.7 &1.15 &  58.0 \\
\textcolor{gray}{\textit{Closed-source model}}  \\
GPT-4-turbo (Summary Conc.) & \xmark &84.6&1.08&62.7&0.79&54.5&0.88&66.1 \\
GPT-4-turbo (Direct Conc.) & \xmark  &84.3  &1.42  &62.3  & 0.82 &54.2 &0.64 &65.9  \\
\textcolor{gray}{\textit{Large reasoning model}}  \\
DeepSeek-R1 (Summary Conc.) & \xmark &\colorbox{secondBest}{85.6}  &0.93  &\colorbox{firstBest}{\textbf{64.3}} & 0.69 &\colorbox{secondBest}{56.5} &1.05 &\colorbox{secondBest}{65.8}  \\
DeepSeek-R1 (Direct Conc.) & \xmark & \colorbox{firstBest}{\textbf{86.4}} & 0.79 & 63.3 &0.77  &\colorbox{firstBest}{\textbf{58.7}} &0.83 &\colorbox{firstBest}{\textbf{67.3}}  \\
       \midrule
    \multicolumn{9}{c}{{\textit{Synthetic Dataset (Total 115 Questions)}}} \\
    \midrule
      \textcolor{gray}{\textit{Open-source model: Llama3.1-8B}} \\
    Knowledge Editing & \cmark &76.2  &0.67  &47.3  &0.83  &27.4 &1.23 & 34.0 \\
    LoRA-Tune & \cmark &77.7  &0.89  &51.2  &0.93  &31.2 & 1.07&  40.7 \\
    Summary Concatenation & \xmark &83.3  &0.79  &52.7  &1.07  &46.6 &0.97 &  50.2 \\
    Direct Concatenation & \xmark &83.6  & 0.83 & 61.4 &1.25  &45.2 & 1.24&6.70 \\
\textcolor{gray}{\textit{Closed-source model}}  \\

GPT-4-turbo (Summary Conc.) & \xmark & 84.2 &0.91  &74.5  &0.72  &61.1 &0.95 &73.3  \\
GPT-4-turbo (Direct Conc.) & \xmark & 85.4 & 0.76 & \colorbox{firstBest}{\textbf{75.5}} &  0.69&\colorbox{secondBest}{62.9} &0.89 &\colorbox{firstBest}{\textbf{75.6}}  \\
\textcolor{gray}{\textit{Large reasoning model}}  \\
DeepSeek-R1 (Summary Conc.) & \xmark &\colorbox{secondBest}{85.7}  &0.92  &70.1  &0.87  &62.7 &0.93 &73.5  \\
DeepSeek-R1 (Direct Conc.) & \xmark & \colorbox{firstBest}{\textbf{87.6}} &0.93  &\colorbox{secondBest}{74.7}  &0.94  &\colorbox{firstBest}{\textbf{67.4}} &0.88 &\colorbox{secondBest}{74.2}  \\

    \bottomrule
    \end{tabular}}
 \end{table*}

\section{Evaluation}

\label{sec:evaluation}

\subsection{Experimental Setup}
\label{subsec:experimental_setup}

\paragraph{Evaluation Methods.}  
To assess the model's ability to retain and utilize knowledge, we design a comprehensive evaluation framework. When posing questions about the current episode \(E_t\), all preceding episodes \(E_1\) to \(E_{t-1}\), including dialogues, locations, and temporal information, serve as contextual knowledge. The evaluation methods fall into two broad categories: (\romannumeral1) parametric and (\romannumeral2) non-parametric approaches.  

(\romannumeral1) Parametric methods involve modifying the model's internal representations to enhance knowledge retention. One such approach is Knowledge Editing-Grace~\cite{DBLP:conf/nips/HartvigsenSPKG23} , which directly alters the model’s weights to integrate new knowledge while preserving existing capabilities. Another technique, LoRA Fine-Tuning~\cite{DBLP:conf/iclr/HuSWALWWC22}, employs low-rank adaptation to efficiently update parameters. This method is computationally lightweight and mitigates catastrophic forgetting, making it particularly effective for incremental learning.  

(\romannumeral2) Non-parametric methods, in contrast, rely on external mechanisms to manage contextual information. Direct Concatenation maintains information integrity by appending historical context directly to the input. While this prevents information loss, its effectiveness is constrained by the model’s context window size. To address this limitation, Summary Concatenation leverages GPT’s abstraction capabilities to extract and condense key information from past episodes. This approach balances information compression with retention, making it a practical solution for handling extensive context.

\paragraph{Model Selection.}  
 We selected the most recent and widely adopted models as our backbone architectures, encompassing open-source model (Llama3.1-8B~\cite{meta2024llama3}), closed-source models (GPT-4-turbo~\cite{DBLP:journals/corr/abs-2303-08774}), and state-of-the-art reasoning model (DeepSeek R1~\cite{DBLP:journals/corr/abs-2303-08774}). The distinguishing characteristics of these models are presented in Table~\ref{tab:model_comparison}.

\subsection{Experimental Results}
\label{subsec:experimental_results}

\paragraph{Evaluation Protocol.}  
We follow the LLM-as-Judge paradigm~\cite{DBLP:conf/nips/ZhengC00WZL0LXZ23}, using the DeepSeek evaluator~\cite{DBLP:journals/corr/abs-2405-04434} for automatic scoring. Each question is paired with a ground truth answer containing factual details and structured reasoning. We use pairwise grading between the model output and ground truth, scoring from 1 to 100. By grounding the evaluation in factual reference answers, this setup ensures more reliable results than open-ended assessments that depend on the model's internal knowledge.

\paragraph{Overall Performance.} 

The results show clear performance differences across models and datasets. Large reasoning models like DeepSeek-R1 and the proprietary GPT-4-turbo outperform the open-source Llama3.1-8B in all tasks. DeepSeek-R1 achieved the highest overall accuracy (67.3\%) on the Hamlet dataset using the direct concatenation method, especially in self-awareness (86.4\%) and relation shift (58.7\%). On the synthetic dataset, GPT-4-turbo also using direct connection achieved the best overall accuracy (75.6\%) and factual memory score (75.5\%).

Non-tuning methods (direct and summary connection) perform better than tuning-based methods (knowledge editing and LoRA-Tune), suggesting that leveraging the model’s original context is more effective, and this is intuitive. All methods perform better on the synthetic dataset than on Hamlet, likely due to its more complex characters, plots, and longer dialog samples (As shown in Table~\ref{tab:dataset_comparison}).

All methods show relatively low standard deviations (most between 0.7-1.2), indicating stable and reliable results. GPT-4-turbo has a higher standard deviation in self-awareness (1.42 on the Hamlet dataset), suggesting some fluctuation. In contrast, DeepSeek-R1 demonstrates more consistent performance, especially in factual memory, with a standard deviation between 0.69-0.94.
Overall, DeepSeek-R1 offers the most balanced performance, excelling in complex relation shift tasks, while GPT-4-turbo excels in factual memory.

\begin{figure*}[t!]
\centering
    \includegraphics[width=\linewidth]{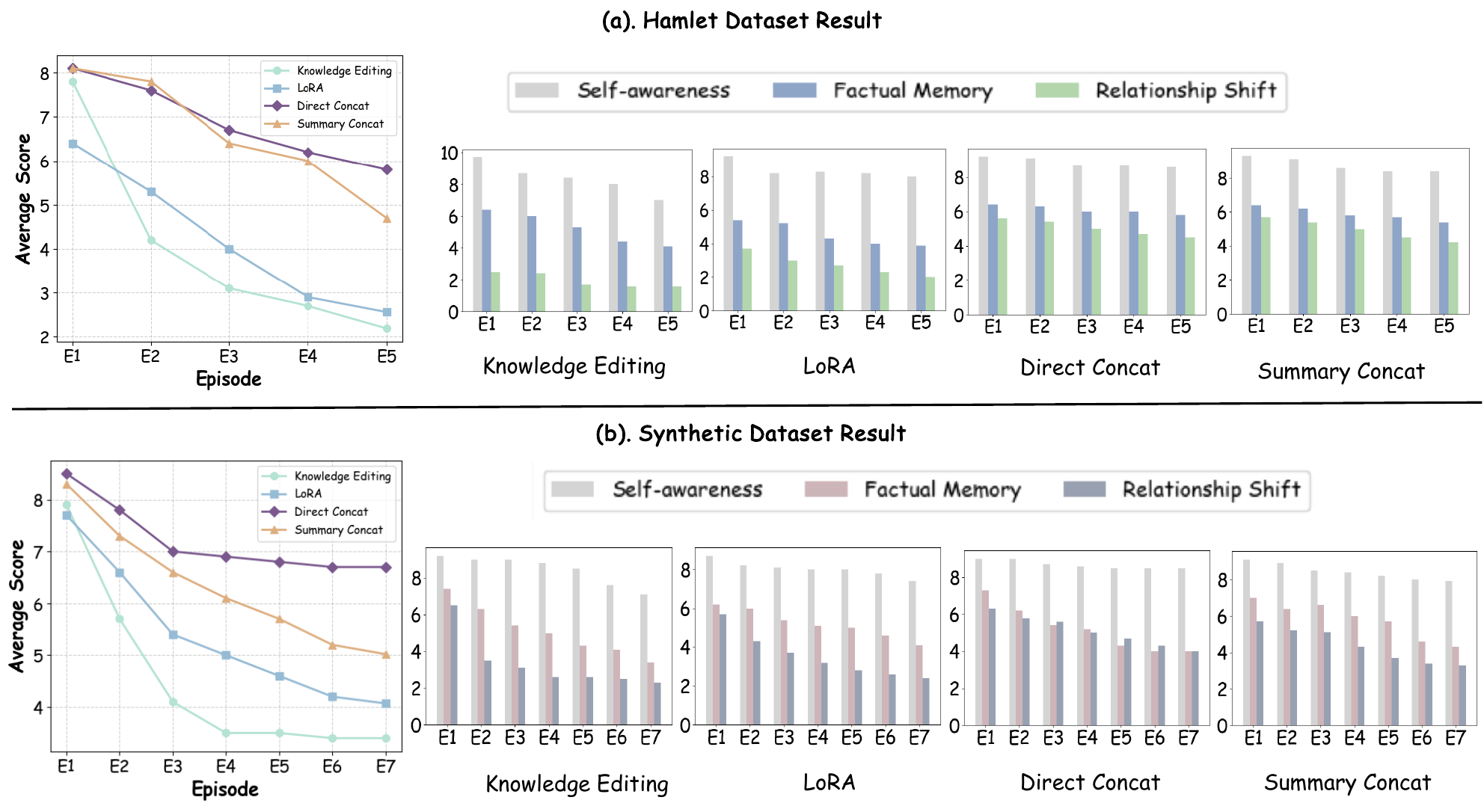}
\caption{Episode-wise Performance of Hamlet and Synthetic Datasets. This includes the overall performance of various methods, as well as performance from different state perspectives.}
\label{fig: episode_trend}
\end{figure*}

\paragraph{Episode-wise Performance.}
Using Llama3.1-8B as an example, we analyzed how each method performs across episodes. As shown in the figure~\ref{fig: episode_trend}, on the Hamlet dataset, model performance generally drops as the story progresses, regardless of parameter tuning. The decline is most severe for the Knowledge Editing method, showing clear signs of catastrophic forgetting. A similar trend appears in the synthetic dataset, suggesting that our \modelname presents challenges for lifelong learning evaluation.

\paragraph{State Dimension Breakdown.}
When broken down by question type, all methods show performance drops over episodes. The most challenging are questions about shifting relationships, where models struggle to track evolving dynamics.

The direct concatenation method performs consistently across question types and datasets. It is especially accurate in early episodes (E1–E2) when handling self-awareness and relationship shift. The summary-concatenation works well for self-awareness and fact recall but performs poorly on relationship shift questions. This suggests it fails to capture complex relationship changes. Knowledge Editing (GRACE) and LoRA-Tune perform weakly on self-awareness and memory-related tasks. Their scores drop quickly over episodes, further showing that parameter-based methods are vulnerable to forgetting in multi-step and long-term reasoning.

\paragraph{Dataset Comparison.} In our case observation, some Hamlet outputs suggest data leakage, even after replacing character names. For example, models sometimes predict future plot details. In contrast, the synthetic dataset avoids such contamination. Yet, models show only slight improvement in relationship shift. This confirms that the main challenge lies in model limitations, not data bias.

We also find that question format matters. Open-ended questions often lead to long and repetitive answers. Structured factual questions improve both accuracy and conciseness, making them better for fair evaluation. This highlights the importance of question design in benchmark construction.

\section{Conclusion}
We introduce \modelname, a novel benchmark designed to evaluate the lifelong learning ability of LLMs through multi-agent, multi-turn interactions. Unlike prior static assessments, \modelname simulates cumulative experiences by organizing interactions as episodic scripts enriched with scene and character dynamics. It enables objective measurement of state evolution via fact-based questions, exploring self-awareness, factual memory retrieve, and relationship shifts. Our experiments on both open-/closed-source and state-of-the-art reasoning models reveal that LLMs still struggle with consistent state retention across episodes. \modelname proves effective in highlighting these challenges and shows that non-parametric methods better preserve long-term context. These results confirm its value as a diagnostic tool for developing more stateful, memory-capable LLMs.

\section{Limitations} While individual samples in the dataset are sufficiently long, the overall number of samples is limited, potentially restricting the diversity of training and evaluation scenarios. Second, the Hamlet dataset may suffer from data contamination, although we have mitigated this issue through name replacement. In the future, we plan to synthesize additional datasets to further enhance the benchmark's robustness.

\section*{Acknowledgments}

This paper was supported by the National Key R\&D Program of China 2024YFE0111800, and NSFC U25B2049.

\bibliography{custom}


\end{document}